# Joint Estimation of Expertise and Reward From Human Demonstrations

Pamela Carreno-Medrano[1]    Stephen L. Smith[2]    Dana Kulić[1]

*Abstract*—When a robot learns from human examples, most approaches assume that the human partner provides examples of optimal behavior. However, there are applications in which the robot learns from non-expert humans. We argue that the robot should learn not only about the human's objectives, but also about their expertise level. The robot could then leverage this joint information to reduce or increase the frequency at which it provides assistance to its human's partner or be more cautious when learning new skills from novice users. Similarly, by taking into account the human's expertise, the robot would also be able of inferring a human's true objectives even when the human's fails to properly demonstrate these objectives due to a lack of expertise. In this paper, we propose to jointly infer the expertise level and objective function of a human given observations of their (possibly) non-optimal demonstrations. Two inference approaches are proposed. In the first approach, inference is done over a finite, discrete set of possible objective functions and expertise levels. In the second approach, the robot optimizes over the space of all possible hypotheses and finds the objective function and expertise level that best explain the observed human behavior. We demonstrate our proposed approaches both in simulation and with real user data.

## I. INTRODUCTION

As the number of robotic systems that collaborate with and assist humans keeps increasing, there is a need for developing methods that allow these systems to quickly learn how to behave in a desired manner.

In order to understand the human actions and leverage this information when planning for an appropriate response, a common approach is for the robot to observe examples of human behavior to estimate the human's objectives when completing a task [1], [2]. Once the robot has a good understanding of the human's objective function, it can imitate the human's demonstration [3], choose actions that will improve collaboration [4], perform tasks according to the human's preferences [5], [6] or coordinate its behavior with other agents or humans [7].

Most of these applications have been developed under an assumption of optimality on the human's demonstrated behavior. That is, the demonstrations the robot observes and uses to estimate the human's objectives are generated by a human that is an expert at completing that given task. However, there are situations in which the robot might need to interact with, assist or learn from non-expert users. Thus, the robot should learn not only about the human's objectives, but also about their expertise level. By knowing the expertise level of the human, the robot can, for example, increase or reduce the frequency of its assistance or decide to cautiously use the examples from a non-expert demonstrator when learning about a new skill.

In this work, we define expertise as "the capability of making action choices that best align with an agent's internal objective" [8]. Thus, a human that frequently chooses more highly valued actions with respect to their internal objective is considered more of an expert than a human that often chooses poorly.

We propose to jointly infer the expertise level and objective function of a human given observations of their (possibly) non-optimal behaviour during a specific task. To do so, we leverage the Maximum Entropy policy framework [9]. This class of policies model stochastic behavior and can capture multiple modes of non-expert behavior (e.g., although the human acts according to the same objective they do not execute the same sequence of actions every time) [10]. Furthermore, they include a temperature parameter $\beta$ that controls the level of stochasticity in the policy and hence provides a suitable approximation to the expertise levels we want to estimate.

We extend our prior Bayesian inference framework for estimating expertise, introduced in [11], to the joint inference of a human's objective function and expertise level. To solve this inference objective, two approaches are proposed. In the first approach, inference is done over a finite, discrete set of possible objective functions and expertise levels. This formulation allows for a quick and easy estimate of the robot's belief over the human's parameters (i.e., their objective and expertise level). However, when using this approach, the robot runs the risk of encountering a human whose behavior is not well explained by the elements of the set, which can result in poor inference results. To address these cases, we propose a second approach in which the robot optimizes over the continuous space of possible hypotheses on human expertise and objectives and finds the objective function and expertise level that best explain the observed human behavior. Finally, we illustrate the utility of the proposed approaches in simulation and with real user data. Our simulation results indicate that the proposed joint inference approach results in performance metrics that are in average 2 orders of magnitude smaller than the one obtained in the cases in which (incorrect) assumptions about a human's true preferences and expertise level are made during inference.

This research is partially supported by the Natural Sciences and Engineering Research Council of Canada (NSERC).

[1]Pamela Carreno-Medrano and Dana Kulić are with the Faculty of Engineering, Monash University, Melbourne, Australia `{Dana.Kulic, Pamela.Carreno}@monash.edu`

[2]Stephen L. Smith is with the Department of Electrical and Computer Engineering, Faculty of Engineering, University of Waterloo, Ontario, Canada `Stephen.Smith@uwaterloo.ca`

The following section reviews the existing work on the estimation of a human's objective function and expertise level. We formally introduce the problem of joint inference of objective functions and expertise levels in Section III. Section IV describes the proposed approaches. In Section V we present the simulation experiments used to validate the proposed approaches. The simulation results for each of the proposed approaches are reported and analyzed in Sections VI and VII. Section VIII illustrates the application of the proposed approach on real user data. We conclude the paper in Section IX with a discussion of our findings, limitations and future work.

## II. RELATED WORK

In this section we provide an overview of the existing approaches for the inference of a human's objective. We then review prior work on the estimation of expertise. We end this section with approaches that address the joint estimation of these parameters.

*Inference of Human Objectives*

The inference of human objectives is one of the most active research topics in the HRI community. Existing approaches mainly differ in the way in which the objective functions are modelled, the type of observed human behavior used during inference and the manner in which this information is collected.

Regarding the modelling of human objectives, existing methods can be categorized into two groups. The first group consists of approaches where human objectives are formulated as a weighted sum of a set of predefined, hand-coded feature functions (e.g., [12], [5]). While this assumption on the structure of the human objective function makes inference tractable and easy to scale, it may be over simplistic and not sufficiently expressive when applied to more complex tasks [13]. The second group considers less structured representations such as deep neural networks [14] or non-linear formulations such Gaussian Processes [15]. This increase in expressiveness comes at the cost of an extensive search space for which a large number of human observations might be needed during inference.

The types of human behavior used for inferring human objectives have substantially expanded in recent years. Jeon et al. [16] recently proposed an unifying formalism for the inference of human objectives under which all types of human behavior are captured. According to this formalism, human behavior can be seen as a choice from an implicit set of options that is approximately rational for the human's intended objective. Demonstrations (e.g., [5] and [12]), corrections (e.g., [5]) and comparisons (e.g., [6], [17], [14]) are all examples of behaviors to be considered during inference. Independently of the type of behavior used during inference, the majority of the existing approaches strongly rely on the assumption that the human from which the robot receives information is an expert. Hence, the demonstrations, corrections or comparisons they provide to the robot truly reflect their internal objective. This dependency on high-quality, expert feedback can potentially hinder the performance of the inference approach used by the robot in the cases in which the observed humans provide poorly executed examples because of they lack of expertise [18].

Although they do not explicitly estimate the expertise level of the human, some recent approaches have considered scenarios in which the human might choose incorrectly when answering two comparison queries [19], they are more likely to answer incorrectly when the alternatives presented by the robot are very similar [12] or they provide ranked sub-optimal demonstrations [18].

*Estimation of a User's Expertise Level*

In the context of Human-Robot Interaction, existing approaches for determining the expertise of a human interacting with the robot can be divided into two main groups. In the first group, supervised learning and task-specific measures of performance are used directly to predict expertise. For instance, Enayati *et al.* [20] trained a binary classifier to distinguish between experts and novices based on seven carefully selected performance metrics. In [21], Hidden Markov Models (HMM) were used to define an expert model that encodes expert surgical gestures. The expertise level of a new user is determined by comparing their performance against this reference model. In the second group, task modeling approaches such as Markov Decision Processes (MDP) are commonly used. Milliken et al. [22] used a hand-tuned partially observable MDP (POMDP) model to predict the likelihood of a user being an expert after each observed task execution. In [11], the rationality coefficient in a softmax Boltzman policy is used to determine a user's expertise level at a simulated kitting task. Most of these approaches require prior domain knowledge about the task or the task objective against which all users are compared. Furthermore, they assume that there is a unique mode of expert behavior and thus can potentially label humans with different action strategies as novices. Compared to Boltzman policies in which only the best way to perform a task is accounted for, the chosen Maximum Entropy formulation allow us to consider all the possible ways in which such a task can be accomplished [23]. This is important because the same human objectives can be achieved in different ways.

The prediction of expertise is also highly relevant in the domain of intelligent learning systems. Two main approaches are commonly used to determine whether a student has mastered, i.e., has a high expertise in, a particular skill [24]. Bayesian knowledge tracing models skills as latent variables in a HMM. The state of these latent variables is updated based on the correctness of the student's answers. Item response theory methods determine whether the probability of skill mastery is conditionally maximized by a given sequence of observed student actions or answers. Both approaches require an expert to define the relations between the performance indicators, skills and actions.

There are few examples in the literature of simultaneous estimation of human objectives and expertise levels. To the best of our knowledge, the work done in [8] is the closest to the ideas proposed in this paper. The authors jointly estimated the expertise level and the motivation coefficient of students during an educational game. However, their method requires observations from the entire student population in order to determine the relative expertise level of each student.

## III. PROBLEM STATEMENT

We consider the setting where a robot $\mathcal{R}$ passively observes a human $\mathcal{H}$ completing a task. The human acts according to their internal policy $\pi_{\mathcal{H}}$. This policy captures the decision process of the human, i.e., how likely the human is to act in a certain manner given the current state of the task.

Following the principle of *rational behavior* [25], we hypothesize that humans are more likely to choose actions that best align with their inner preferences and benefit the pursuit of their goals (i.e., in our case the human's goal is to complete a given task). However, if they have a low expertise level, mistakes (e.g., poor action choices with respect to their preferences and goals) are also more likely to be observed. Thus the policy $\pi_{\mathcal{H}}$ simultaneously encodes the human's preferences when completing a given task as well as their level of expertise.

The robot observes the human complete one or more instances of a task and leverages this information to reason about the human's policy and level of expertise.

### A. Preliminaries

We formulate the inference of the human's policy $\pi_{\mathcal{H}}$ as an inverse reinforcement learning problem in the space of finite-horizon Markov Decision Processes (MDP). An MDP consists of the tuple $(S, A, T, \gamma, R)$, where $S$ and $A$ denote the set of possible task states and human actions respectively, $T : S \times \mathcal{A} \times S \to [0, 1]$ captures the state transition dynamics, that is, the conditional probability distribution over the next state $s^{t+1}$ given the current state of the task $s^t \in S$ and human action $a^t \in A$. The function $R : S \times A \times S \to \mathbb{R}$ assigns a real-value reward to each task state transition and $\gamma$ is a discount factor that determines the relative importance between immediate and future rewards. To simplify notation hereinafter we will abbreviate $R(s^t, a^t, s^{t+1})$ as $R^t$.

A stationary decision policy $\pi$ in an MDP is a mapping $\pi : S \times A \to [0, 1]$ that specifies the probability of selecting an action $a^t$ at a given state $s^t$, with $\sum_{a^t \in A} \pi(s^t, a^t) = 1$. In this work, we approximate the human's policy $\pi_{\mathcal{H}}$ using the maximum entropy (MaxEnt) reinforcement learning formulation [9]:

$$\pi^{\text{MaxEnt}}(s^t, a^t) = \arg\max_{\pi} \sum_t \mathbb{E}_{(s^t, a^t) \sim \pi} \gamma^t [R^t + \beta H(\pi(s^t, \cdot))], \quad (1)$$

in which the expected sum of rewards and the entropy $H((\pi(s^t, \cdot)) = -\sum_{a^t} \pi(a^t | s^t) \log \pi(a^t | s^t)$ at each visited state are maximized. The temperature parameter $\beta$ controls the stochasticity of the resulting policy. That is, as $\beta \to 0$, a conventional deterministic policy in which the actions that maximize the expected reward are always chosen can be recovered from $\pi_{\text{MaxEnt}}$. On the contrary, as $\beta \to \infty$, a policy in which all actions have a non-zero probability of being sampled will be obtained. Our choice of this type of policy formulation is motivated by the fact it *i.)* models stochastic behavior; *ii.)* captures multiple modes of non-expert behavior (e.g., although the human follows the same strategy they do not execute the same sequence of actions every time); and iii.) can be learned from non-optimal observations [9], [10].

### B. Problem Formulation

From Eq. (1) and under the assumption that the task space dynamics, $T(s^t, a^t, s^{t+1})$, are known to the robot, in order to reason about the human's policy $\pi_{\mathcal{H}}$, the robot needs to infer the reward function $R^t$ and the parameter $\beta$. While the former encodes the preferences and objective of the human, the latter relates to the human's expertise level.

Since the robot has no direct access to $R^t$, the robot reasons about approximations of the human's true reward function. We consider parametric approximations of the form

$$\hat{R}(s^t, a^t, s^{t+1}) = \theta^T \phi(s^t, a^t, s^{t+1}), \quad (2)$$

where $\phi : S \times A \times S \to \mathbb{R}^d$ denotes a vector of features over state transitions, which are known to the robot, and $\theta \in \mathbb{R}^d$ with $\theta_i \in [0, \theta^{\max}]$ corresponds to the human's preferences on these state features.

Thus, given observations of the human's behavior $\Xi = \{\xi_1, \xi_2, ..., \xi_k\}$ with $\xi_i = \{(s^0, a^0), (s^1, a^1), \ldots, (s^T, a^T)\}$ during a task, the robot reasons over parameters of the human's reward function $\theta$ and their expertise level $\beta$,

$$P(\theta, \beta | \Xi) = \frac{P(\Xi | \theta, \beta) P(\theta, \beta)}{\int_{\bar{\theta}, \bar{\beta}} P(\Xi | \bar{\theta}, \bar{\beta}) P(\bar{\theta}) P(\bar{\beta}) d\bar{\theta} d\bar{\beta}}, \quad (3)$$

where $P(\Xi | \theta, \beta)$ characterizes how the robot expects the human's actions to be informed by their preferences and expertise level, $\int_{\bar{\theta}, \bar{\beta}} P(\Xi | \bar{\theta}, \bar{\beta}) P(\bar{\theta}) P(\bar{\beta}) d\bar{\theta} d\bar{\beta}$ defines the appropriate normalization constant, and $\bar{\theta} \in \Theta$ and $\bar{\beta} \in \mathcal{B}$ indicate the spaces of all human's preferences and expertise levels.

### C. Human Observation Model

Under the Maximum Entropy (MaxEnt) formulation, the soft Q-function $Q^{\text{soft}}(s^t | a^t)$ describes the expected sum of discounted future rewards and entropy after taking action $a^t$ in state $s^t$ and following a policy $\pi(s^t, a^t)$

$$Q^{\text{soft}}(s^t, a^t) = \hat{R}^t + \mathbb{E}_{s^{t+1}, a^{t+1} \ldots \sim \pi} \left[ \sum_{l=1}^{\infty} \gamma^l (\hat{R}^{t+l} + \beta H(\pi(s^{t+l}, \cdot))) \right]. \quad (4)$$

Similarly, the value function $V^{\text{soft}}(s^t)$ describes the expected sum of discounted future rewards and entropy when starting at state $s^t$ and thereafter following the same policy $\pi(s^t, a^t)$

$$V^{\text{soft}}(s^t) = \beta \log \left[ \sum_{a \in A(s^t)} \exp\left(\frac{1}{\beta} Q^{\text{soft}}(s^t, a)\right) \right]. \quad (5)$$

Haarnoja et al. [26] have shown that the policy that optimizes the MaxEnt objective in Eq. (1) is given by

$$\pi_{\text{MaxEnt}}(s^t, a^t) = P(a^t|s^t) = \exp\left[\frac{1}{\beta}\left(Q^{\text{soft}}(s^t|a^t) - V^{\text{soft}}(s^t)\right)\right]. \tag{6}$$

Thus, given a human's preferences $\theta$ and expertise level $\beta$, Eq. (6), the probability of observing the human take an action $a^t$ in state $s^t$ can be rewritten as

$$\pi^{\text{MaxEnt}}_{\theta,\beta}(s^t, a^t) = P_{\theta,\beta}(a^t|s^t) = \exp\left[\left(\frac{1}{\beta}(Q^{\text{soft}}_{\theta,\beta}(a^t|s^t)\right.\right. \tag{7}$$
$$\left.\left. - V^{\text{soft}}_{\theta,\beta}(s^t)\right)\right],$$

where both $V^{\text{soft}}_{\theta,\beta}(s^t)$ and $Q^{\text{soft}}_{\theta,\beta}(s^t|a^t)$ can be obtained through the following fixed-point iteration updates

$$Q^{\text{soft}}(s^t, a^t) \leftarrow \hat{R}^t + \gamma \mathbb{E}_{s^{t+1} \sim T(\cdot|s^t, a^t)}\left[V^{\text{soft}}(s^{t+1})\right] \forall s^t, a^t; \tag{8}$$

$$V^{\text{soft}}(s^t) \leftarrow \beta \log\left[\sum_{a^t \in A(s^t)} \exp\left(\frac{1}{\beta}Q^{\text{soft}}(s^t, a^t)\right)\right] \forall s^t. \tag{9}$$

Finally, using Eq. (7), the likelihood of the robot observing a demonstration $\xi = \{(s^0, a^0), (s^1, a^1), \ldots, (s^T, a^T)\}$ generated by a human with preferences $\theta$ and expertise $\beta$ can be defined as

$$P(\xi|\theta,\beta) = \prod_{(s^t,a^t) \in \xi} \exp\left[\left(\frac{1}{\beta}(Q^{\text{soft}}_{\theta,\beta}(a^t|s^t) - V^{\text{soft}}_{\theta,\beta}(s^t))\right)\right]. \tag{10}$$

In the case of multiple observation sequences $\Xi = \{\xi_1, \xi_2, \ldots, \xi_k\}$, Eq. (10) can be extended as follows

$$P(\Xi|\theta,\beta) = \prod_k \prod_{(s^t,a^t) \in \xi_i} \exp\left[\left(\frac{1}{\beta}(Q^{\text{soft}}_{\theta,\beta}(a^t|s^t) - V^{\text{soft}}_{\theta,\beta}(s^t))\right)\right]. \tag{11}$$

## IV. PROPOSED APPROACH

The robot maintains a joint Bayesian belief over the human's parameters, $b(\theta,\beta)$. The robot starts with a non-informative uniform prior over the continuous space of possible hypotheses on human expertise and preferences. After observing a new sequence $\xi$ of the human's behavior during a given task, the robot updates its belief[1] according to

$$b'(\theta,\beta) = \frac{P(\xi|\theta,\beta)b(\theta,\beta)}{\int_{\bar{\theta},\bar{\beta}} P(\xi|\bar{\theta},\bar{\beta})b(\bar{\theta},\bar{\beta})d\bar{\theta}d\bar{\beta}}, \tag{12}$$

where $P(\xi|\theta,\beta)$ corresponds to the likelihood defined in Eq. (10), $b(\theta,\beta)$ is the robot's belief prior to observing $\xi$, and $b'(\theta,\beta|) = P(\theta,\beta|\Xi)$ as per Eq. (3) with $\Xi = \{\xi\}$.

[1]We note that this Bayesian update can also be done every time a new action is observed. This modification is useful when few sequences of human behavior are available during inference.

Two approaches are proposed for the joint inference of the human's preferences $\theta$ and their expertise level $\beta$. In the first approach, hereinafter referred to as *discrete set*, the robot reasons over a finite, fixed set of possible hypotheses when updating its belief over $b(\theta,\beta)$. In the second approach, hereinafter referred to as *optimization-based*, the robot seeks to determine the values $(\hat{\theta}, \hat{\beta})$ that maximize Eq. (12).

### A. Discrete Set Inference

Given discretized and finite spaces of $\beta \in \mathcal{B}_D$ and $\theta \in \Theta_D$, we define $\Psi = \{(\theta,\beta) | \theta \in \Theta_D, \beta \in \mathcal{B}_D\}$ as the set of possible human hypotheses considered by the robot during inference. After each new observed trajectory of the human's behavior $\xi = \{(s^0, a^0), (s^1, a^1), \ldots, (s^T, a^T)\}$, from Eq. (12), the robot updates its belief over the set of all hypotheses according to

$$b'(\theta,\beta) = \frac{P(\xi|\theta,\beta)b(\theta,\beta)}{\sum_{\bar{\theta} \in \Theta, \bar{\beta} \in \mathcal{B}} P(\xi|\bar{\theta},\bar{\beta})b(\bar{\theta},\bar{\beta})}, \forall (\theta,\beta) \in \Psi, \tag{13}$$

where $P(\xi|\theta,\beta)$ is the observation model introduced in Eq. (10). The merit of the *discrete set* approach lies in the fact that this normalization constant, and hence, the posterior $b(\theta,\beta)$ can be quickly and easily computed by summing over the discrete hypothesis set.

### B. Optimisation-Based Inference

When reasoning about the human's parameters $\theta$ and expertise level $\beta$ using a finite discrete set of hypotheses $\Psi$, the robot runs the risk of encountering a human user whose behavior is not well explained by the models in the set. To address this limitation, we formulate an approach to determine point estimates of the human's parameters $\theta$ and $\beta$ from the observed human behavior. Using Maximum a Posteriori (MAP) optimization, we aim to find the $\hat{\theta}$ and $\hat{\beta}$ parameters that maximize the posterior distribution introduced in Eq. (3) given all observed trajectories $\Xi$,

$$\begin{aligned} \max_{\hat{\theta},\hat{\beta}} \quad & P(\Xi|\theta,\beta)P(\theta,\beta) \\ \text{subject to} \quad & \|\hat{\theta}\|_1 \leq \lambda, \\ & \hat{\beta} > 0, \end{aligned} \tag{14}$$

where $\lambda$ is a pre-specified free parameter that determines the amount of sparsity in the human's preferences $\hat{\theta}$ and $P(\theta,\beta)$ encodes any prior knowledge the robot has about the human's preferences and expertise level.

Since the exact posterior is complex and non-trivial to implement in a tractable manner, we approximate it using an adapted version of the Monte Carlo Markov Chain sampling algorithm proposed in [27]. In this algorithm, the posterior distribution is sampled by placing a Markov Chain at the intersection points of a grid of length $\delta$ in the region $[0, \theta_{\max}]^{|\theta|+1}$. At each iteration, the sampler moves to a neighboring point and uses the current estimate of the MDP's policy, $Q^{\text{soft}}$ and $V^{\text{soft}}$ functions to sample the posterior distribution at that point. The original method has been adapted to jointly optimize for $\theta$, $\beta$ using MaxEnt policies and the appropriate definitions of the policy evaluation and policy iteration algorithms as described in [26], [28]. Currently, the optimization is run offline using a set of trajectories of observed human behavior. We note also that contrary to most inverse reinforcement learning approaches in which observations are assumed to be nearly optimal with respect to the unknown parameters $\theta$, since we aim at jointly inferring a human's expertise level and preferences, the proposed *optimization-based* approach considers observations that are possibly far from optimal.

---

**Algorithm 1:** Extended BIRL

**Input** : MDP M\R, step sizes $\delta$ and $\lambda$, set of demonstrations $\Xi$, max. number of iterations $K$, prior distribution $P(\theta, \beta)$
**Output:** Maximum a posteriori estimates $\hat{\theta}, \hat{\beta}$

Pick a random preferences vector $\hat{\theta} \in [0, \theta_{\max}]^{|\theta|}$
Pick a random expertise level $\hat{\beta} \sim \text{Unif}(0.01, \beta_{\max})$
Compute $Q^{\text{soft}}_{\hat{\theta},\hat{\beta}}$ and $V^{\text{soft}}_{\hat{\theta},\hat{\beta}}$ as per Eq. (8) and Eq. (9)
**for** $i \leftarrow 1$ **to** $K$ **do**
  Pick a preferences vector $\widetilde{\theta}$ uniformly at random from the neighbours of $\hat{\theta} \in [0, \theta_{\max}]^{|\theta|}/\delta$
  Pick an expertise level $\widetilde{\beta}$ uniformly at random from the neighbours of $\hat{\beta} \in [0, \beta_{\max}]/\lambda$
  Compute $Q^{\text{soft}}_{\widetilde{\theta},\widetilde{\beta}}$ and $V^{\text{soft}}_{\widetilde{\theta},\widetilde{\beta}}$ using soft-policy evaluation as defined in [28]
  Compute $P(\hat{\theta}, \hat{\beta}|\Xi) \leftarrow P(\Xi|\hat{\theta}, \hat{\beta}) P(\hat{\theta}, \hat{\beta})$ as per Eq. (11)
  Compute $P(\widetilde{\theta}, \widetilde{\beta}|\Xi) \leftarrow P(\Xi|\widetilde{\theta}, \widetilde{\beta}) P(\widetilde{\theta}, \widetilde{\beta})$ as per Eq. (11)
  Sample $a \sim \text{Uniform}(0,1)$
  **if** $a < \min\left\{1, \frac{P(\widetilde{\theta},\widetilde{\beta}|\Xi)}{P(\hat{\theta},\hat{\beta}|\Xi)}\right\}$ **then**
    Set $\hat{\theta} \leftarrow \widetilde{\theta}$ and $\hat{\beta} \leftarrow \widetilde{\beta}$
    Set $Q^{\text{soft}}_{\hat{\theta},\hat{\beta}} \leftarrow Q^{\text{soft}}_{\widetilde{\theta},\widetilde{\beta}}$, and $V^{\text{soft}}_{\hat{\theta},\hat{\beta}} \leftarrow V^{\text{soft}}_{\widetilde{\theta},\widetilde{\beta}}$
  **end**
**end**

---

## V. EVALUATION IN SIMULATION

We first test the robot's capability of jointly inferring a human's preferences and expertise level in simulation using a variety of randomly generated environments. The aims of our simulations are three-fold: *i.)* to empirically demonstrate the need for a joint inference of a human's preferences and expertise level, *ii.)* to test the correctness of the proposed inference approaches, and *iii.)* to objectively assess the benefit of inferring both the human preferences $\theta$ and expertise $\beta$ compared to the case in which $\theta$ is fixed for all humans and only their expertise coefficient is inferred as done in [11]. In the following we detail the simulation environment, task and human preference model used for this evaluation.

### A. Environment

Similar to [6] and [29], we consider a scenario where a mobile robot has to traverse a 2D environment containing a finite number of user defined zones. Each zone corresponds to specific traversal rules or restrictions the robot should take into consideration when moving within the environment. We refer to these user defined zones as the environment specification. When moving within the environment, the robot can choose from four discrete actions that deterministically move the robot one step in each of the cardinal directions.

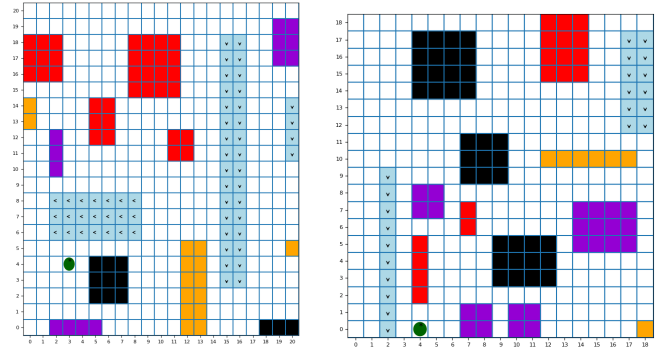

Fig. 1: Examples of randomly generated simulation environments. Each color corresponds to a different traversal zone. Red indicates zones to be avoided, black are obstacles, yellow indicates zones with speed limits, purple indicates zones with high circulation and blue indicates roads with preferred directions of traversal. The green dot indicates the goal state.

Our experiments were performed on 5 different environments, each generated in a semi-random manner. To generate all environments, we first specified 5 types of user-defined zones: areas to be avoided (red), preferred roads with specific traversal directions (light blue), slow down areas (yellow) and spaces with high traffic (purple) and static obstacles (black). Next, we randomly sampled the size of the environment, the number of zones (up to 4) of each type to be included as well as their locations and dimensions. Examples of some generated environments are shown in Fig. 1.

### B. Task

We assume humans have preferences over how strictly the robot should follow the rules associated with each zone and the relative importance among zones. To inform the robot about these preferences, a human provides demonstrations of tele-operated traversals to a given goal destination (green dot in Fig. 1) from multiple randomly sampled initial locations. However, since humans have different levels of expertise when controlling the mobile robot, the demonstrated traversals are potentially sub-optimal with respect to their preferences. Thus, when inferring information from the human's observed traversals, the robot should simultaneously reason about the human's preferences and expertise level.

### C. Human Preferences and State Features

Given an environment specification, five state features are considered: path length ($\phi_1$) and binary values that indicate whether the current position of the robot in the grid falls within a road ($\phi_2$), a zone to avoid ($\phi_3$), a slow-down zone ($\phi_4$) or high traffic area ($\phi_5$). To enforce traversal directions and avoidance of restricted areas, a penalty is applied every time the tele-operated robot traverses a road in a direction contrary ($\phi_2$) to the one specified and enters or exits a restricted area ($\phi_3$ or $\phi_5$). Similarly, to encourage the usage of preferred roads as well as shorter paths, a reward is given every time the robot transits along a road and once the target location is reached. Formally, the reward function of a human tele-operating the robot in a simulated environment is defined as follows

$$R^t = \begin{bmatrix} -\theta_1 \\ \theta_2 \\ \theta_3 \\ \theta_4 \\ \theta_5 \end{bmatrix} \begin{bmatrix} \phi_1 & \mathbf{1}_{\phi_2} & \mathbf{1}_{\phi_3} & \mathbf{1}_{\phi_4} & \mathbf{1}_{\phi_5} \end{bmatrix} \quad (15)$$

where $\mathbf{1}_{\phi_i}$ is the indicator function associated with the $i^{th}$ state feature and $t$ indicates the current time-step in the tele-operation trajectory. This function takes on a positive or negative value depending on the reward or penalty associated to each zone in the user specification. Similarly, when the current state does not belong to a user specified zone, the indicator functions are set to 0.

Given this definition of the state features, the preferences vector $\theta$ captures: *i.)* the relative importance attributed to following the path specifications associated with each type of zone, and *ii.)* the trade-off between shorter paths with a (potentially) high number of restriction violations and longer paths in which specifications are followed more closely.

### D. Hypotheses Set For Discrete Approach

To define the hypotheses set $\Psi$ considered during the discrete inference approach, we followed a discretization approach similar to the one proposed in [5]. The set $\Theta$ was first restricted to the space of positive vectors with unit $l_1$-norm. Each component $\theta_i \in \theta$ was allowed to take values $0.0, 0.3, 0.5, 0.7,$ or $1.0$. Since $\theta \in \mathbf{R}^5$, $\Theta$ is equivalent to the 5-fold Cartesian product with cardinality $|\Theta| = 3124$. The set of possible expertise levels was defined as $\mathcal{B} = \{0.01, 0.09, 0.5, 1.0, 5.0, 10\}$. These values allow us to evaluate the performance of the proposed approaches when the stochasticity of the simulated humans' policies is high ($\beta \in \{5, 10\}$), optimal and non-optimal choices (i.e. mistakes) are approximately equally probable ($\beta \in \{0.5, 1.0\}$), and the optimal actions are more frequently taken ($\beta \in \{0.01, 0.05\}$). Given our definition of expertise, these three cases correspond to simulated humans with low, medium and high expertise respectively.

Since multiple levels of expertise are considered during inference, if all vectors in $\Theta$ were to be used, the set $\Psi$ will include more than 15000 hypotheses. To reduce the size of the hypothesis set, we sampled a set $\Theta_k$ of $k$ preference vectors, with $k \in \{5, 10, 15, 20\}$ from $\Theta$. By doing so, we can assess how advantageous it is to jointly infer a human's preferences and expertise level while simultaneously analyzing how the discrete set approach scales to the number of hypotheses considered during inference. Finally, we denote by $\Psi_k = \{(\theta, \beta) | \theta \in \Theta_k, \beta \in \mathcal{B}\}$ the set of hypothesis in which $k$ distinct preferences vectors are considered.

### E. Performance Metrics

Three performance metrics are used to evaluated the proposed joint inference formulation. We first define $\theta^*$ and $\beta^*$ as the true preferences and expertise level the robot is trying to infer from the human's demonstrated behaviour. Similarly, we denote by $\hat{\theta}$ and $\hat{\beta}$ the preferences and expertise estimate obtained by the robot during inference.

*Expertise Metric:* To measure the accuracy on the estimation of a human's true expertise level, we compute the absolute distance between the true value $\beta^*$ and the estimated value $\hat{\beta} = \int_{\Theta, \mathcal{B}} \beta P(\theta, \beta | \Xi) d\beta d\theta$. The larger this difference, the poorer the performance of the inference method.

*Preferences Metric:* To measure the accuracy on the estimation of a human's preferences, we use the cosine similarity between the true preferences $\theta^*$ and the estimated vector $\hat{\theta} = \int_{\Theta, \mathcal{B}} \theta P(\theta, \beta | \Xi) d\theta d\beta$. The cosine similarity is defined as follows

$$\text{cosine\_similarity}(\theta^*, \hat{\theta}) = 1 - \frac{\theta^* \cdot \hat{\theta}}{||\theta^*||^2 ||\hat{\theta}||^2} \quad (16)$$

The closer the metric is to 1, the greater the similarity between the true and predicted human preferences.

*Policy Regret:* Previous work [30] has found that different preferences vectors might still lead to the same observed behavior. In the context of our experiments, this means that preferences estimates $\hat{\theta}$ whose cosine similarity with respect to the true human's preferences is low can result in behaviours that are close to the demonstrations obtained for the true preferences vectors $\theta^*$. The policy regret metric accounts for this fact. It compares the true state values $V^{\text{soft}}_{\theta^*, \beta^*}(s) \forall s \in S$ to the ones obtained using the estimated preferences vectors $V^{\text{soft}}_{\hat{\theta}, \beta^*}(s)$ and provides us with an approximate measure of the similarity between the action choices and behaviors (as stated in Eq. 7) induced by the estimated and true human preferences. The closer the regret value is to 0, the better the estimate. We formally define regret as follows

$$\left| V^{\text{soft}}_{\theta^*, \beta^*}(\cdot) - V^{\text{soft}}_{\hat{\theta}, \beta^*}(\cdot) \right| = \frac{1}{|S|} \sum_{s \in S} \frac{\left| V^{\text{soft}}_{\theta^*, \beta^*}(s) - V^{\text{soft}}_{\hat{\theta}, \beta^*}(s) \right|}{\max \left( V^{\text{soft}}_{\theta^*, \beta^*}(\cdot) \right)}. \quad (17)$$

Since the magnitude of the state values also depends on the expertise levels $\beta^*$, we normalize the state value differences by the maximum value of true value function $V^{\text{soft}}_{\theta^*, \beta^*}(\cdot)$ so as to make the regret metrics obtained for all expertise levels comparable.

## VI. RESULTS - BENEFITS OF A JOINT INFERENCE

In the following we demonstrate the performance and benefits of the proposed joint inference approach. The results presented in this section were obtained using the *discrete set* method previously described in Sec. IV-A. This method allows for an easy and quick approximation of the posterior probability distribution $P(\theta, \beta | \Xi)$ and thus facilitates the comparison with approaches in which the humans' preferences or expertise levels are assumed to fixed or known ahead of time.

First, in Sec. VI-B we show that an inference approach for the estimation of expertise in which the human's preference is (incorrectly) assumed to be known results in estimates that frequently over or undervalue the true expertise of the observed humans. Second, in Sec. ?? and VI-C, we demonstrate that preference estimates obtained under assumptions of known expertise are more likely to be incorrect, in particular when observations are obtained from non-experts. Third, in Sec. VI-E, we analyze the effect of the number of episodes, the number of hypotheses considered during inference and the complexity of the MDP environment on the performance of the discrete set approach. Finally, in Sec. VI-E, we examine the robustness of the discrete set approach. As expected, the performance of the discrete set approach degrades when the true parameters are not included in the set of hypotheses considered during inference and this decrease in performance correlates to how dissimilar the true parameters are from the set.

### A. Evaluation Conditions

For each one of the 5 randomly generated environments first described in Sec. V-A, the *discrete set* inference was run 5 times. For each run and for each hypotheses set $\Psi_k$ with $k \in \{5, 10, 15, 20\}$, three inference instances, one for each expertise level, i.e., low, medium, and high, were tested. For each instance, the true parameters $(\theta^*, \beta^*)$ were randomly chosen from the hypotheses set $\Psi$ and a total of 20 episodes, i.e., same target location with randomly sampled initial states, were generated using these parameters. A total of 300 (5 environments × 5 runs × 4 hypotheses sets $\Psi_k$ × 3 expertise levels) instances were executed. For each instance, three inference conditions were examined:

- **Full Set**, where all parameters in the hypotheses set are considered during inference with $\theta^* \in \Theta$ and $\beta^* \in \mathcal{B}$. This condition corresponds to the joint inference formulation proposed in this work.
- **Fixed $\theta$ Set**, in which the set of hypotheses is redefined as $\Theta = \{\theta\}$ with $\theta \neq \theta^*$ and $\beta^* \in \mathcal{B}$. This condition evaluates whether a human's expertise level can be inferred when assuming (incorrectly) that their preferences are known.
- **Fixed $\beta$ Set**, where the set of hypotheses is redefined $\mathcal{B} = \{\beta\}$ with $\beta \neq \beta^*$ and $\theta^* \in \Theta$. This condition gauges whether a human's preferences can be inferred when a wrong assumption about their true expertise level is made during inference. In this condition, $\beta$ is chosen to belong to an opposite expertise group than $\beta^*$. For instance, if $\beta^*$ corresponds to a high expertise, $\beta$ is set to a low or medium expertise value.

Finally, to evaluate the performance of the discrete approach in the cases in which the true human parameters are not part of the hypotheses set, a second set of values $(\theta^*_{\text{out}}, \beta^*_{\text{out}})$, hereinafter referred as the *out of set model*, was also selected. The true expertise level $\beta^*_{\text{out}}$ was uniformly sampled from the [0.01, 10] interval and the true preference vector $\theta^*_{\text{out}}$ was randomly selected from the $\Theta \setminus \Theta_k$ set.

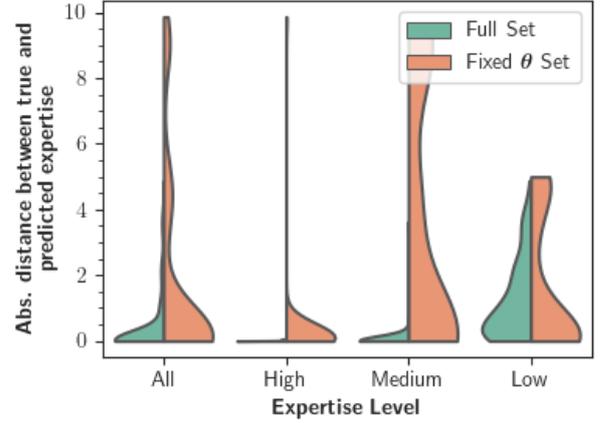

Fig. 2: Distribution of the absolute distance between true expertise values $\beta^*$ and the estimates obtained using joint inference (**Full Set** in green) and a fixed preference vector for all simulated humans (**Fixed $\theta$ Set** in orange) by expertise level. Results are shown for all runs, environments and hypotheses sets combined

### B. Condition One: Fixed Preferences Set

This condition compares the expertise estimates obtained using the proposed joint inference formulation, i.e., **Full Set**, to the case in which all humans are assumed to share the same preferences and reward function, i.e., **Fixed $\theta$ Set**, as done in [11]. As our metric of performance, we measured the absolute distance between the true value $\beta^*$ and the expected values $\hat{\beta}_{\text{FullSet}} = \sum_{i=1}^{|\mathcal{B}|} \beta_i P_{\mathcal{B}}(\beta_i)$ and $\hat{\beta}_{\text{Fixed}\theta} = \sum_{i=1}^{|\mathcal{B}|} \beta_i P(\beta_i)$, where $P_{\mathcal{B}}(\beta)$ denotes the marginal distribution $P_{\mathcal{B}}(\beta) = \sum_{\mathcal{B}} b(\theta, \beta)$. The expected values $\hat{\beta}_{\text{FullSet}}$ and $\hat{\beta}_{\text{Fixed}\theta}$ were obtained after all 20 episodes were seen during inference. Fig. 2 depicts the distribution of the absolute distance between true and predicted expertise levels for all environments, runs and hypotheses sets $\Psi_k$ combined. Results are presented for all expertise levels.

Two main findings are observed: First, the absolute distance between the expertise estimates obtained from the **Full Set** $\hat{\beta}_{\text{FullSet}}$ and the true expertise level $\beta^*$ is smaller than the distance between the true value $\beta^*$ and the estimates obtained under the assumption of fixed preferences $\hat{\beta}_{\text{Fixed}\theta}$. This difference is more apparent in the high and medium expertise instances. Second, for the cases of low expertise, we found that both approaches were able to determine that the observations were generated by simulated humans with low expertise, but failed to distinguish between the two possible expertise values in this group, i.e., $\beta^* \in \{5.0, 10.0\}$. We hypothesize that in order to better distinguish between these values more observations are needed.

To test this hypothesis, an additional set of 100 inference instances were executed (5 environments × 5 runs × 4 hypotheses sets $\Psi_k$ × 1 expertise level) with a maximum number of 40 episodes. Results are presented in Fig. 3. Overall, the expertise estimates obtained using the proposed joint inference approach improve with the number of episodes seen during inference. No improvement in predicted expertise levels was observed for the **Fixed $\theta$ Set** approach.

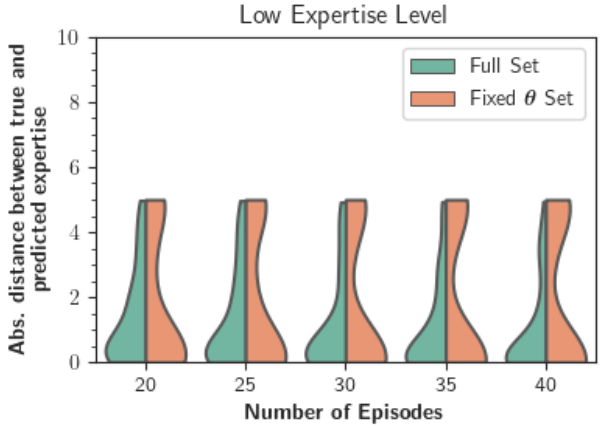
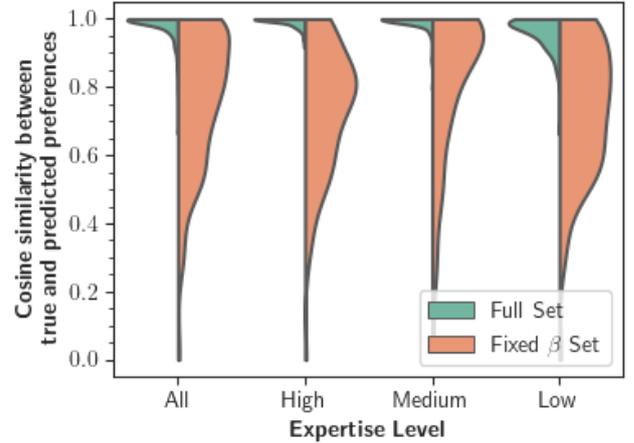

Fig. 3: Distribution of the absolute distance between true expertise values $\beta^*$ and the estimates obtained using joint inference (**Full Set** in green) and a fixed preference vector for all simulated humans (**Fixed $\theta$ Set** in orange) for the low expertise case. Results are shown by the number of episodes seen during inference with all runs, environments and hypotheses sets combined

Fig. 4: Distribution of cosine similarity metrics between the estimated preference vectors $\hat{\boldsymbol{\theta}}_{\text{FullSet}}$ (green), $\hat{\boldsymbol{\theta}}_{\text{Fixed}\beta}$ (orange) and the true value $\boldsymbol{\theta}^*$ for fixed expertise condition. The closer to 1 the the stronger the similarity between true and predicted preferences

### C. Condition Two: Fixed Expertise Set

This condition compares the preference estimates obtained using our joint inference approach, i.e., **Full Set**, to the case in which all simulated humans are assumed to have the same expertise level, i.e., **Fixed $\beta$ Set**. The latter case is frequently found in the learning preferences literature [27], [31] where observed demonstrations are assumed to come from human experts.

To gain a better understanding on the effects of different expertise assumptions, the fixed expertise values were set to belong to an expertise group opposite to the true level of the simulated human from which observations were generated. For instance, if the simulated human had a high expertise, the expertise level used in the **Fixed $\beta$ Set** was chosen from the low or medium expertise groups. Comparisons were made based on the expected values $\hat{\boldsymbol{\theta}}_{\text{FullSet}} = \sum_{i=1}^{|\Theta|} \boldsymbol{\theta}_i P_\Theta(\boldsymbol{\theta}_i)$ and $\hat{\boldsymbol{\theta}}_{\text{Fixed}\beta} = \sum_{i=1}^{|\Theta|} \boldsymbol{\theta}_i P(\boldsymbol{\theta}_i)$, where $P_\Theta(\boldsymbol{\theta})$ denotes the marginal distribution $P_\Theta(\boldsymbol{\theta}) = \sum_\Theta b(\boldsymbol{\theta}, \beta)$. Both estimates were obtained after a total of 20 episodes were seen during inference. As an initial metric of performance, we measured the cosine similarity between the true human's preferences $\boldsymbol{\theta}^*$ and the estimates obtained using both approaches $\hat{\boldsymbol{\theta}}_{\text{FullSet}}$ and $\hat{\boldsymbol{\theta}}_{\text{Fixed}\beta}$. Fig. 4 depicts the distribution of the cosine similarity metric between true and predicted preferences vectors for all environments, runs and hypotheses sets $\Psi_k$. Results are presented for all expertise levels (see Fig. 4).

The spread and concentration of values in the leftmost violin in Fig. 4 indicate that, across all expertise levels, our proposed **Full Set** approach was able to predict estimates closer to the true preferences. As in the **Fixed $\theta$** condition, we observe that the **Full Set** approach performs best when observations came from simulated humans with true high and medium expertise levels. The decrease in similarity observed for the **Full Set** approach in the low expertise case might be explained by the fact that different preferences can potentially lead to the same observed behavior as demonstrated in [30]. Since in the case in low expertise, random actions are more likely to happen, it is possible that the same random action was observed for numerically different preferences vectors. To test this hypothesis, we measured the policy regret the true state values and the state values obtained using each estimate according to Eq. 17.

We note that to make the state values comparable, $V^{\text{soft}}_{\hat{\boldsymbol{\theta}}_{\text{FullSet}}, \beta^*}(\cdot)$ and $V^{\text{soft}}_{\hat{\boldsymbol{\theta}}_{\text{Fixed}\beta}, \beta^*}(\cdot)$ were computed using the true expertise level and not the fixed value used during inference. As in the case of the cosine similarity metric, results were combined for all environment, runs and hypotheses sets $\Psi_k$. Results are presented for all expertise levels.

Overall we observe that as initially indicated by the cosine similarity comparison, the **Full Set** approach performs better for all expertise levels than the **Fixed $\beta$ Set**. Despite the decrease in cosine similarity previously identified in the low expertise case, regret measures for this case are near to 0. This suggests that the state values and the action choices induced by these state values for the $\hat{\boldsymbol{\theta}}_{\text{FullSet}}$ are similar to the ones observed for the true preferences vectors $\boldsymbol{\theta}^*$.

The preferences estimates obtained under erroneous assumptions of expertise (i.e., **Fixed $\beta$ Set**) result in regret metrics that are approximately 2 orders of magnitude larger than the regret metrics obtained using joint inference (i.e, **Full Set**). This observation is noteworthy because in the existing literature it is often assumed that observed demonstrations come from humans with high expertise [31]. However, we observe that when this initial assumption is not met (in reality the human demonstrator has a lower expertise), inferring preferences under an assumption of high expertise (or optimality) can result in estimates with higher regret. Hence, robots should jointly reason about the human's preferences and expertise level.

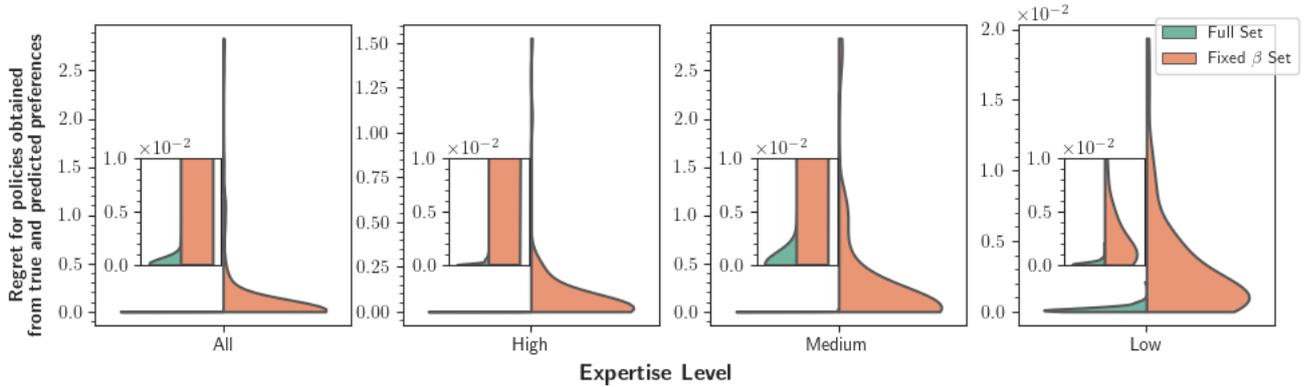

Fig. 5: Distribution of policy regret, that is, the difference between the value of the MaxEnt policy that maximizes the inferred human preferences, $\hat{\theta}_{\text{FullSet}}$ (green) and $\hat{\theta}_{\text{Fixed}\beta}$ (orange), and the value of the MaxEnt policy that maximizes the true human preferences vector $\theta^*$. The closer to 0 the stronger the similarity between the policies obtained from the true and predicted human preferences.

### D. Condition Three: True Model is Excluded from Hypotheses Set

To evaluate the performance of the discrete set approach when the demonstrations were generated by a simulated human whose true expertise level and preferences are not part of the set, we compared all performance metrics, i.e., policy regret, preferences cosine similarity, and absolute distance between true and predicted expertise values, for the selected values $(\theta^*_{\text{in}}, \beta^*_{\text{in}})$ and $(\theta^*_{\text{out}}, \beta^*_{\text{out}})^2$. Comparisons were made based on the expected values $\hat{\theta}_{\text{in}|\text{out}} = \sum_{i=1}^{|\Theta|} \theta_i P_\Theta(\theta_i)$ and $\hat{\beta}_{\text{in}|\text{out}} = \sum_{i=1}^{|\mathcal{B}|} \beta_i P_\mathcal{B}(\beta_i)$ where $P_\Theta(\theta)$ and $P_\mathcal{B}(\beta)$ denotes the marginal distributions $P_\Theta(\theta) = \sum_\Theta b(\theta, \beta)$ and $P_\mathcal{B}(\beta) = \sum_\mathcal{B} b(\theta, \beta)$ respectively. Both estimates were obtained after the maximum number of episodes available (i.e., 20 episodes) were seen during inference.

Performance metrics obtained for both cases and are summarized in Fig. 6. Results are presented for the hypotheses set $\Psi_k$ with cardinality $k = 20$ with all expertise levels and environments combined. Three main patterns were observed. First, there is a decrease in the cosine similarity (middle plot) between the true $\theta^*_{\text{out}}$ and predicted $\hat{\theta}_{\text{out}}$ preferences when compared to the cases in which the true preferences were included in the hypotheses set. Second, the average distance between the true values $\beta^*_{\text{out}}$ and their respective estimates $\hat{\beta}_{\text{out}}$ triples in comparison to the cases in which the true expertise level is part of the hypotheses set, i.e., $\beta^*_{\text{in}}$ and their respective estimates $\hat{\beta}_{\text{in}}$. Third, regret metrics approximately 6 times larger were observed for the *out of set* instances compared to the cases in which the true parameters are part of the hypotheses set.

In order to gain further insight into how the composition of the hypotheses set affects the estimates obtained for the *out of set* test cases, we computed and analyzed Pearson correlation coefficients ($\rho$) between all performance metrics and (a) the average cosine similarity between $\theta^*_{\text{out}}$ and $\Theta_k$, (b) the diversity of the hypotheses set as measured by the average cosine similarity among all elements in the set, and (c) the minimum distance between the true expertise level $\beta^*_{\text{out}}$ and all elements in $\mathcal{B}$. Results are listed in TABLE I.

[2] The subscripts in and out are used to indicate whether the true preferences and expertise level are included in or excluded from the discrete hypotheses set respectively.

| Factor | Preferences Similarity | Expertise Distance | Regret |
|---|---|---|---|
| | **Metrics** | | |
| Similarity between $\theta^*_{\text{out}}$ and $\Theta_k$ | 0.57† | −0.45† | 0.05 |
| Similarity within $\Theta_k$ | −0.06 | −0.03 | −0.12 |
| Min. distance between $\beta^*_{\text{out}}$ and $\mathcal{B}$ | 0.0001 | 0.58† | −0.33† |

TABLE I: Paired *Pearson* correlation coefficients ($\rho$) between all performance metrics and the composition of the discrete hypotheses set as measured by (a) the average cosine similarity between $\theta^*_{\text{out}}$ and $\Theta_k$, (b) the average cosine similarity among all elements in the set, and (c) the minimum distance between the true expertise level $\beta^*_{\text{out}}$ and the elements in $\mathcal{B}$. Statistical significance is indicated with this symbol †.

Two main trends with statistical significance ($p$−value < 0.05) are identified. First, as indicated in the first row in TABLE I, both the cosine similarity ($\rho = 0.57$) and the absolute distance between the predicted an true expertise levels ($\rho = −0.45$) metrics improve when the similarity between $\theta^*_{\text{out}}$ and $\Theta_k$ increases. Second, regarding the expertise estimates, we observe that the further apart $\beta^*_{\text{out}}$ is from the elements in $\mathcal{B}$ (see third row in TABLE I), the larger the distance between the predicted and true expertise values ($\rho = 0.58$). Similarly, we observed that the further $\beta^*_{\text{out}}$ is from the elements in $\mathcal{B}$, the larger the impact that wrongly inferred parameters have on the regret for the resulting policy ($\rho = −0.33$).

Thus, as expected, the overall performance of the discrete set approach degrades when the true parameters are not included in the set of hypotheses considered during inference, and the performance decrease is correlated to how far the true human preferences and expertise level are from the elements included in the hypotheses set.

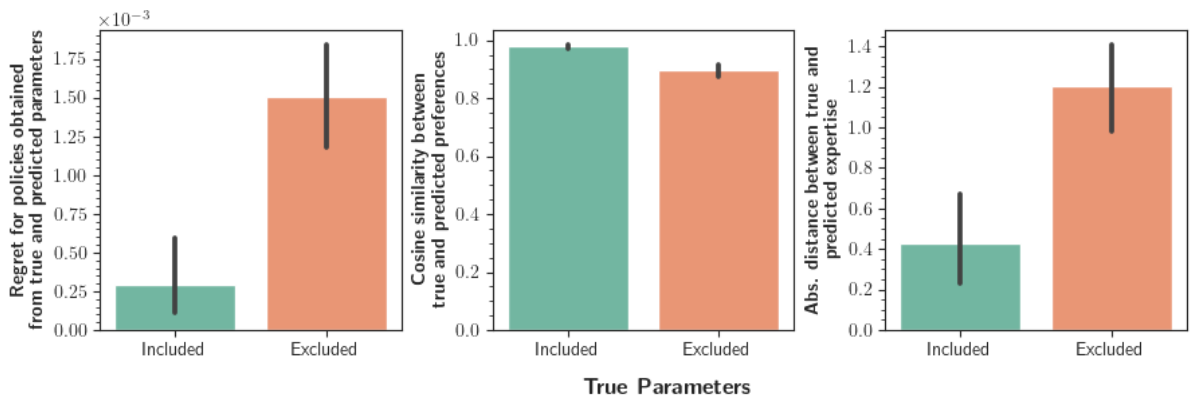

Fig. 6: Average performance metrics across environments (i.e., policy regret, cosine similarity between true and predicted human preferences and abs. distance between true and predicted expertise) observed for discrete set approach when true values are included (In Set) or exclude (Out Set) as hypotheses. Error bars indicate 95% confidence intervals. Metrics were computed using the preferences and expertise estimates obtained after 20 episodes were seen during inference.

*E. Sensitivity Analysis*

The previous section showed that the discrete set inference method, although easy and fast to compute, is limited by the diversity of hypotheses considered during inference. In this section we further analyse how the performance of this inference method is influenced by other factors such as the number of episodes, cardinality of the hypotheses set, and environment complexity (as measured by the number of states). To do so, we have computed *Pearson* correlation coefficients ($\rho$) between these factors and the performance metrics obtained for the *in set* instances. That is, the inference cases in which the true expertise level and human's preferences were part of the hypotheses set. Correlation coefficients for all expertise levels combined are listed in TABLE. II.

Significant correlation coefficients (*p*-value < 0.05) were found between the number of episodes and all performance metrics and the number of models and the cosine similarity and regret metrics. In particular, we observe that as the number of episodes seen during inference increases, the cosine similarity between the true and predicted preferences increases, and regret and the distance between true and predicted expertise level decreases. Fig 7 shows the changes observed in all performance metrics, for all expertise levels, as a function of the number of episodes seen during inference.

With regards to the correlation between the number of episodes and all performance metrics, three main patterns are identified. First, we observe that expertise estimates require fewer observations than the preferences estimates. By the time at least 5 episodes have been seen, the discrete set inference method was able to determine the correct expertise group for most instances (see range[3] of absolute distance values in Fig 7 left plot). The remaining episodes allowed to differentiate among the possible expertise values within the same expertise group as indicated by the continuous improvement of the distance metrics in the medium and low expertise cases. Second, for the preferences estimates, we observed that a good performance is obtained after seeing at least 15 episodes during inference for all expertise levels and the improvement observed at the 20 episodes mark is well within the margin-of-error seen after 15 episodes. Hence, the effect of those additional 5 episodes is small. Third and last, although both the expertise distance and cosine similarity metrics kept improving as more episodes were seen during inference, policy regret values near 0 were observed for all expertise levels around the 10 episode mark. This indicates that the proposed joint inference formulation was able to produce expertise and preference estimates that resulted in a similar decision policy as the one associated to the true human parameters $\theta^*$ and $\beta^*$.

[3]Maximum distance values for each expertise group: 0.08 (high), 0.5 (medium) and 5.0 (low)

|  | Metrics | | |
|---|---|---|---|
| Factor | Preferences Similarity | Expertise Distance | Regret |
| **No. Episodes** | 0.39† | −0.26† | −0.27† |
| **Cardinality of $\Psi_k$** | −0.09† | 0.02 | 0.06† |
| **No. States** | −0.02 | 0.03 | 0.01 |

TABLE II: Paired *Pearson* correlation coefficients ($\rho$) between the number of episodes, the number of models and environment complexity (as defined by the number of states), and all performance metrics. Statistical significance is indicated with this symbol †.

Concerning the influence of the cardinality of the hypotheses set in the performance of the discrete set method, it was observed that both cosine similarity and policy regret degrade as the number of hypotheses considered during inference increases. This might by explained by the fact that numerically dissimilar preferences can produce the same observed trajectories and the likelihood of such cases increases with the cardinality of the hypotheses set.

We further analyze the impact of the cardinality of the hypotheses set by considering additional *in set* and *out of set* inference instances for which the number of candidate preference vectors included in $\Psi_k$ was set to $k = \{20, 40, 60, 80, 100\}$. The performance of the discrete set inference method with these larger hypotheses set was tested in a single simulated environment. *Pearson* correlation coefficients ($\rho$) between all performance metrics and the cardinality of the hypotheses set were obtained for all instances. Results are listed in TABLE. III.

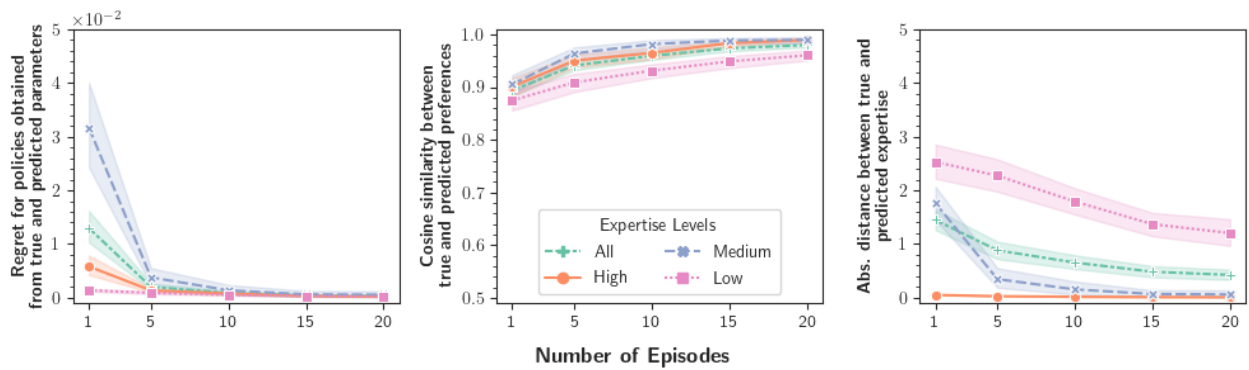

Fig. 7: Average performance metrics (i.e., policy regret, cosine similarity between true and predicted human preferences and abs. distance between true and predicted expertise) as the number episodes seen during inference increases. Results are grouped by expertise level. Bands indicate 95% confidence intervals.

| | Metrics | | |
|---|---|---|---|
| **Inference Case** | **Preferences Similarity** | **Expertise Distance** | **Regret** |
| In set ($\theta^*_{in}\beta^*_{in}$) | 0.03 | 0.04 | 0.14† |
| Out of set ($\theta^*_{out}\beta^*_{out}$) | 0.20† | 0.06 | −0.12† |

TABLE III: Paired *Pearson* correlation coefficients ($\rho$) between cardinality of hypotheses set (i.e., $\Psi_k$ with $k = \{20, 40, 60, 80, 100\}$) and all performance metrics for both *in set* and *out of set* inference instances. Statistical significance is indicated with this symbol †.

Two opposite trends are observed. One the one hand, for the cases in which the true human preferences and expertise level are included in the hypotheses set, we found that the discrete set performance as measured by policy regret ($\rho = 0.14$, $p-$values $< 0.05$) degrades when the number of preferences vector in the hypotheses set increases. On the other hand, when the true human preferences and expertise level are excluded from the hypotheses set, improvements in performance as indicated by the cosine similarity and policy regret metrics were observed when the number of hypotheses considered by the discrete set method increased. Thus, although easy and quick to compute, the discrete set method is highly dependent on which and how many hypotheses are included during inference.

## VII. RESULTS - OPTIMIZATION APPROACH

The discrete set approach is sensitive to the choice of the hypotheses set considered during inference. Thus, the robot is at risk of wrongly estimating a human's expertise level and preferences if the hypothesis set is incomplete. The optimization-based inference method first introduced in Sec. IV-B addresses this problem, since it considers the continuum of all possible values, i.e., $\theta^*_i \in [0, 1]$ and $\beta^* \in (0, \infty)$ during inference.

In the following we evaluate the optimization-based inference method. First, in Sec. VII-B we compare it to the discrete set method. Specifically, we analyze the performance of both inference methods when the true human preferences and expertise level are not part of the discrete set hypotheses. Our results show that the preferences and expertise estimates obtained using the optimization method outperform, across all performance metrics, the estimates obtained using the discrete set method. Second, in Sec. VII-C, we investigate the effect of number of episodes, average episode length and environment complexity on the performance of the optimization approach. Our results indicate that performance increases with the number of episodes and the length of these episodes. Similarly, we found that the complexity of the environment has no effect in the performance of the optimization inference method. Finally, in Sec. VII-D, we compare the computation time required for both approaches. As expected, the discrete set approach is faster to compute compared to the optimization approach, however the latter has better performance.

### A. Evaluation Conditions

A total of 100 inference instances (20 instances × 5 sampled environment) were run using both the optimization and discrete set approaches. For each instance, the true parameters ($\theta^*, \beta^*$) where selected as follows: First, the true expertise level was sampled from the $\beta^* \sim$Unif(0.01, 5) distribution. We decided to limit the set of possible expertise values to the [0.01, 5] interval since, as mentioned in Sec. VI, it was observed that the episodes obtained from simulated humans with $\beta^* = 5$ tend to not differ from the observations generated with expertise levels set to $\beta^* > 5$. Second, the hypotheses in the discrete set approach were selected as described in Sec. V-D with $k = 20$. Third and last, the true preference vector $\theta^*$ was randomly selected from $\Theta \setminus \Theta_k$ where $\Theta$ corresponds to the 5-fold Cartesian product from which the preference vectors included in the discrete hypotheses, i.e., $\Theta_k$, set were chosen.

For each instance, a maximum of 40 episodes, i.e., same target location with randomly sampled initial states, were generated using the selected ($\theta^*, \beta^*$) parameters. Both the discrete set and optimization inference approaches were run 4 times.

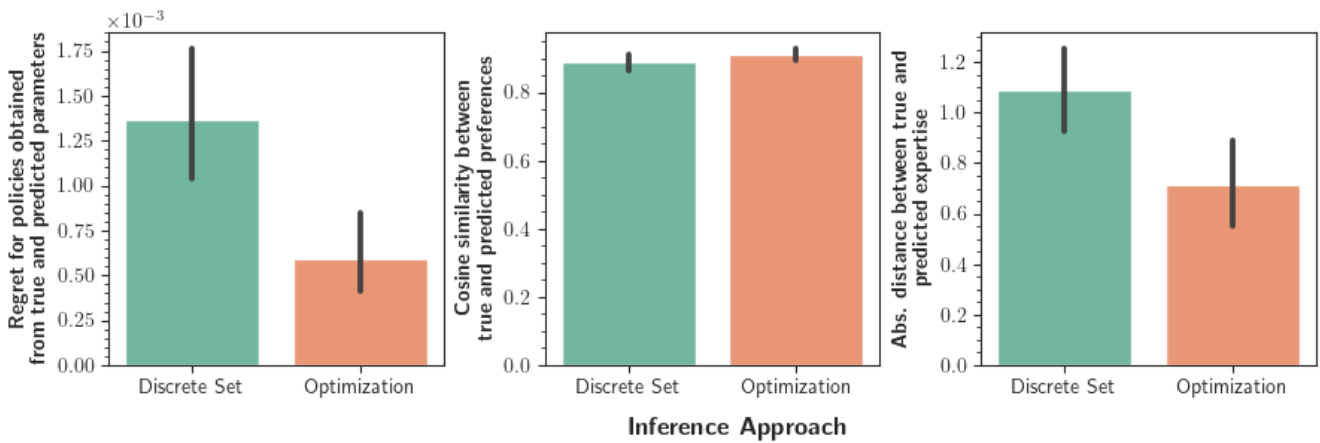

Fig. 8: Average performance metrics (i.e., policy regret, cosine similarity between true and predicted human preferences and abs. distance between true and predicted expertise) observed for both approaches for all environments combined. Error bars indicate 95% confidence intervals. Metrics were computed using the point estimates obtained with a maximum number of observations (40 episodes).

Each time the number of episodes seen during inferences was set to $10, 20, 30$, or $40$. The optimization method was limited to a maximum of 1000 iterations and the priors $P(\theta)$ and $P(\beta)$ were set to a uniform distribution on $[0, 1.0]^{|\theta|}$ for the preference vectors and a normal distribution ($\mu = 5.005$, $\sigma = 1.5$) over the $[0.01, 10.0]$ interval for the expertise level. We note that although the selection of $\beta^*$ was limited to the $[0.01, 5]$ interval, the optimization method was allowed to run in the $[0.01, 10.0]$ interval in order to match the set of expertise values included in the discrete set approach.

### B. Performance Comparison Between Approaches

We consider the same set of performance metrics, first introduced in Sec. V-E. Point estimates obtained for the discrete set approach correspond to the expected values $\hat{\beta}_{\text{Set}} = \sum_{i=1}^{|\mathcal{B}|} \beta_i P_{\mathcal{B}}(\beta_i)$ and $\hat{\theta}_{\text{Set}} = \sum_{i=1}^{|\Theta|} \theta_i P_{\Theta}(\theta_i)$ where $P_{\mathcal{B}}(\beta)$ and $P_{\Theta}(\theta)$ denote the marginal distributions $P_{\mathcal{B}}(\beta) = \sum_{\mathcal{B}} b(\theta, \beta)$ and $P_{\Theta}(\theta) = \sum_{\Theta} b(\theta, \beta)$ respectively. For the optimization approach, the point estimates $\hat{\beta}_{\text{Opt}}$ and $\hat{\theta}_{\text{Opt}}$ correspond to the mean predicted values obtained from the adapted MCMC sampling algorithm used to solve Eq. 14.

Average performance metrics obtained with the maximum number of episodes (i.e. 40 episodes) are presented in Fig. 8 for all environments and expertise levels combined. Overall, we observe that the optimization estimates outperform the estimates obtained using the discrete set approach for all metrics. In particular, we observe that the average absolute distance between $\beta^*$ and $\hat{\beta}_{\text{Set}}$ is approximately 1.5 times larger that the average expertise distance observed between $\beta^*$ and $\hat{\beta}_{\text{Opt}}$ (see rightmost plot in Fig. 8). Although both approaches result in preference estimates that are close to the true preference vector $\theta^*$ as indicated by the middle plot in Fig. 8, we observe that the regret obtained for $\hat{\beta}_{\text{Set}}$ and $\hat{\theta}_{\text{Set}}$ is approximately 2.5 times larger than the regret metrics obtained for the optimization approach estimates $\hat{\beta}_{\text{Opt}}$ and $\hat{\theta}_{\text{Opt}}$. The proximity between the cosine similarity metrics obtained for both methods (i.e, an average cosine similarity of 0.88 and 0.90 for $\hat{\theta}_{\text{Set}}$ and $\hat{\theta}_{\text{Opt}}$ estimates respectively) might be explained by the similarity between $\theta^*$ and the elements in the discrete hypotheses set. This hypothesis is supported by a statistically significant Pearson correlation coefficient ($\rho = 0.57$, $p$-value $< 0.05$) found for the average cosine similarity between $\theta^*$ and $\Theta_k$ and the similarity observed between $\theta^*$ and $\hat{\theta}_{\text{Set}}$ (see first row in TABLE I).).

### C. Sensitivity Analysis

As done in Sec. VI-E, here we analyze the effect of the number of episodes and environment complexity (as measured by the number of states) in the performance metrics obtained for the optimization approach estimates $\hat{\beta}_{\text{Opt}}$ and $\hat{\theta}_{\text{Opt}}$. Moreover, our previous analysis on the performance of the discrete set method indicated that performance metrics vary across expertise levels. Thus, to gain further insights into how and why performance metrics are influenced by the true expertise levels we are trying to estimate, we extend our analysis to two additional factors: true expertise level and average episode length. The inclusion of episode length is motivated by the fact that higher expertise will result in shorter episodes and fewer observations (i.e., simulated experts frequently choose optimal actions and thus reach the target location in fewer steps than simulated humans with medium or low expertise), which in turn limits the amount of information available to the robot during inference and can result in less accurate estimates for the high expertise cases.

*Pearson* correlation coefficients ($\rho$) between the above-mentioned factors and all performance metrics (i.e., cosine similarity, policy regret and absolute distance between true and predicted expertise levels) were computed and analyzed for all 400 instances (20 instances $\times$ 5 environments $\times$ 4 possible no. of episodes). Results are listed in TABLE. IV. As initially observed in the discrete set method, the number of episodes used during inference was found to be statistically significant for the performance of the optimization estimates. This pattern is depicted in more detail in Fig. 9. Leftmost and right most plots show that, as the number of episodes increased, both policy regret ($\rho = -0.36$) and the absolute distance between

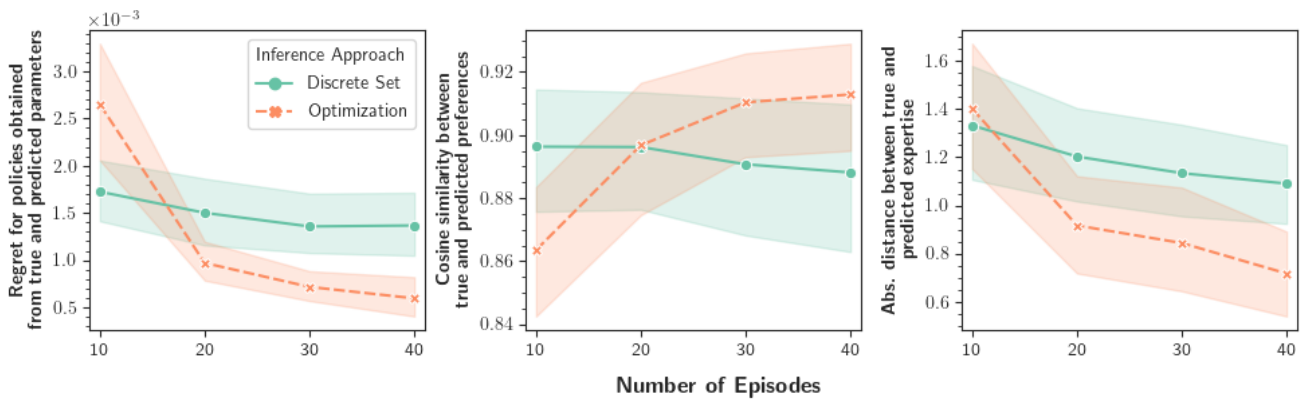

Fig. 9: Average performance metrics (i.e., policy regret, cosine similarity between true and predicted human preferences and abs. distance between true and predicted expertise) of the optimization approach (orange curve) as the number episodes seen during inference increases. Bands indicate 95% confidence intervals. For comparison purposes, metrics for the discrete set approach are also depicted (green curve).

$\boldsymbol{\theta}$ and $\beta^*$ decreased ($\rho = -0.22$). The middle plot illustrates a similar pattern for the cosine similarity between $\hat{\boldsymbol{\theta}}_{\text{Opt}}$ and $\boldsymbol{\theta}^*$; the similarity between the true and predicted preferences improved as the the number of episodes ($\rho = 0.18$) increased.

|  | Metrics | | |
|---|---|---|---|
| **Factor** | **Preferences Similarity** | **Expertise Distance** | **Regret** |
| **No. of episodes** | 0.18† | −0.22† | −0.36† |
| **No. of states** | −0.07 | 0.13† | −0.01 |
| **Average episode length** | −0.12† | 0.31† | −0.13† |
| **True expertise level** | −0.11† | 0.32† | −0.28† |

TABLE IV: Paired *Pearson* correlation coefficients ($\rho$) between the number of episodes, environment complexity (as defined by the number of states), average episode length and true expertise and all performance metrics. Statistical significance is indicated with this symbol †.

To provide a better idea of the significance of the improvement observed for the optimization approach, Fig. 9 also depicts the performance of the discrete set approach as the number of episodes increases. Overall, we observe that as soon as the number of observations used during inference reaches 20 episodes, the performance of the optimization approach is equally good (as shown in the middle plot in Fig. 9) or better than the performance of the discrete set approach (see leftmost and rightmost plot in Fig. 9).

Regarding the other factors, three main patterns are of interest. First, statistically significant positive correlations between the absolute distance between $\hat{\beta}_{\text{Opt}}$ and $\beta^*$ and the true expertise level and average episode length factors were observed. This indicates that as $\beta^* \to \infty$, observed demonstrations tend to be longer (i.e, simulated humans with low expertise take longer to reach the target location), and the accuracy of $\hat{\beta}_{\text{Opt}}$ decreases. A similar decrease in preference similarity was observed for the cases in which $\beta^* \to \infty$. Second, we observed statistically significant negative correlations between policy regret and the true expertise level and the average episode length factors. These correlation coefficients seem to indicate that shorter episodes, which are more likely to happen for instances in which $\beta^* \to 0$, resulted in less accurate estimates and a decrease in the performance of the optimization-based inference method. Finally, a significant correlation between the environment complexity and the absolute distance between $\hat{\beta}_{\text{Opt}}$ and $\beta^*$ was observed. This seems to indicate that as the number of states in the environment increases less accurate estimate expertise are obtained.

### D. Run Time Comparison

We measured and compared the computation time required for each approach when solving for each one the 400 instances used to evaluate the performance of the optimization approach. Results were obtained on a 2.5 GHz E5-2680 v3 processor with 32 GB RAM. We used the Pytorch library [32] to implement all matrix computations required to solve for $\boldsymbol{Q}^{\text{soft}}(s^t, a^t)$ and $V^{\text{soft}}(s^t)$ on K80 GPU with 12 GB of memory.

Overall we observed that the discrete set inference method is on average 4 times faster than the optimization-based method. This is mainly due to the number of times $\boldsymbol{Q}^{\text{soft}}(s^t, a^t)$ must be computed during inference. With the discrete approach, $\boldsymbol{Q}^{\text{soft}}(s^t, a^t)$ is computed just once for every hypothesis in the set. On the contrary, the Monte Carlo Markov Chain sampling method used to solve Eq. 14 requires an update of $\boldsymbol{Q}^{\text{soft}}(s^t, a^t)$ every time that a significant change in $P_{\boldsymbol{\theta},\beta}(s^t|a^t)$ is detected. Future work will focus on achieving better computation times for the optimization approach by including an analytical solution of the gradient in Eq. 14.

## VIII. EVALUATION ON REAL USER DATA

We next apply our proposed approach to a human demonstration dataset. In [29], eight users were provided with a floor plan of a simulated warehouse environment as well as additional contextual information about the various areas within the environment. They were asked to create an environment specification that defines the navigation constraints that a robot transporting material from one location to another should follow in order to safely navigate the environment. The users could specify two types of constraints: *roads*, which indicated direction and a preferred path for the robot, and *restricted* areas where the robot was not allowed to enter or navigate through. Once a user finished with their environment specification, they were asked to tele-operate the robot through the environment until a pre-defined target location was reached. Cartesian trajectories were recorded for each user. We used six of these specifications and trajectories (one trajectory was missing and another had multiple gaps) to demonstrate the proposed method with real human demonstrations.

### A. Setup and Implementation

Each specification was mapped to a discrete environment similar to the one used in our simulations, with $|S| = 4646$. State representation was extended to include both the 2D location of the tele-operated robot as well as its direction (vertical or horizontal). The user trajectories were discretized and down-sampled to one observed state per second. After inspecting the trajectories, it was found that most of the users provided commands to the tele-operated robot at this frequency. The action space was defined to include four discrete actions, i.e., right, left, down and up, that deterministically move the tele-operated robot one step in the specified direction. If the specified action is contrary to the robot's current direction, the robot's direction is adjusted accordingly. For example, if the robot is initially moving in the horizontal direction followed by an "up" or "down" action, its direction will change from horizontal to vertical and vice-versa.

A 3D state feature vector is used to capture the user's preferences during the tele-operation task. The first term, $\phi_1$, corresponds to path length, the second term, $\phi_2$, indicates whether the robot entered or transited the restricted areas in the current state, and the third term, $\phi_3$, specifies whether the robot used a preferred road along the preferred travel direction when transitioning to the current state. The definition of path length was also extended so actions that resulted in a change of direction were twice as costly as those that moved the robot along its current direction, to capture the cost of turning. As done in the simulation study, a penalty factor was added to the respective feature every time the robot entered a restricted area or travelled outside the specified roads. A bonus was also added every time a road was used along the preferred travel direction. Formally, the reward function of a user tele-operating the robot is defined as follows

$$R^t = \begin{bmatrix} -\theta_1 \\ \theta_2 \\ \theta_3 \end{bmatrix} \begin{bmatrix} \phi_1 & \mathbf{1}_{\phi_2} & \mathbf{1}_{\phi_3} \end{bmatrix} \quad (18)$$

where $\mathbf{1}_{\phi_i}$ is the indicator function associated with the $i^{th}$ state feature. This function takes on a negative value if a user specification is violated and is set to 0 when the current state does not belong to a user specified zone. Given this definition, the inferred preference vector $\theta$ shows the relative importance a user associated to reaching the target location as quickly as possible versus to driving according to the specifications they previously defined. Similarly, the inferred expertise level $\beta$ indicates how good a user was at generating a tele-operation trajectory that reflected their true preferences.

Given the limited number of observations (only one execution per user), the discrete inference approach was used to determine a user's preferences. The hypothesis set $\Psi = \{(\theta, \beta) | \theta \in \Theta, \beta \in \mathcal{B}\}$ was defined as follows. First, the set of possible preferences vectors $\Theta$ was restricted to the space of positive vectors with unit $l_1$-norm. Each component $\theta_i \in \theta$ was allowed to take values $0.0, 0.5,$ or $1.0$. After normalization, a total of $|\Theta| = 19$ unique candidate vectors were obtained. Second, the set of possible expertise levels was defined as $\mathcal{B} = \{0.1, 0.3, 1, 3, 10\}$. This discretization suffices to cover high ($\beta \to 0$) and low expertise ($\beta \to \infty$) cases as well as some values in between (i.e, user is not a novice, but still makes some mistakes during tele-operation). We note that in this particular evaluation instance what matters is the relative magnitude of $\beta$ and its ordering relative to other users. A total of $|\Psi| = 95$ hypotheses were evaluated during inference. Finally, since the observations available for inference were limited to a single episode, the Bayesian update first introduced in Eq. 3 was done after every action instead.

### B. Results

Since the true user preferences and expertise levels are unknown in this case, we validate the results obtained from the user data in two different ways. First, we compare general performance metrics such as the total number of actions taken by each user, the total number of times each specification was violated and the number of times preferred roads were traversed to each user's inferred expertise and preferences. Second, we compare the observed users' trajectories to the ones generated using the inferred preferences and a high expertise coefficient (i.e., $\beta = 0.001$). In this manner, we can determine whether the proposed joint inference approach is capable of inferring the user's unknown preferences when the user failed to provide good demonstrations due to a lack of expertise.

For each user, the expertise and preference estimates correspond to the expected values $\hat{\beta} = \sum_{i=1}^{|\mathcal{B}|} \beta_i P_{\mathcal{B}}(\beta_i)$ and $\hat{\theta} = \sum_{i=1}^{|\Theta|} \theta_i P_{\Theta}(\theta_i)$ where $P_{\mathcal{B}}(\beta)$ and $P_{\Theta}(\theta)$ denote the corresponding marginal distribution $P_{\mathcal{B}}(\beta) = \sum_{\mathcal{B}} b(\theta, \beta)$ and $P_{\Theta}(\theta) = \sum_{\Theta} b(\theta, \beta)$, obtained after all their actions were seen during inference. Estimated preferences and expertise levels as well as counts on how many times each zone specification was violated are listed in TABLE V.

*1) Observed Trajectory Analysis:* From the estimated preferences and expertise levels listed in TABLE V and based on the relative expertise ordering among users, we have identified two main user groups. The first group, with predicted expertise levels ($\hat{\beta}$ {0.3}), includes users whose tele-operated trajectories seem to be in closer alignment with their inferred preferences. Overall, the trajectories observed for the users belonging to this group (i.e., users 01, 02, and 06) showed a good trade-off between reaching the target destination and respecting most of the user's specifications (approximately 40% to 60% importance was attributed to reaching the goal state as quickly as possible and 36% to 60% importance was assigned to following their specifications). The main difference between the users belonging to this group is the use of and importance attributed to the preferred roads and travel directions. While users 01 and 06 were mainly concerned with the avoidance of restricted areas and hence only travelled along the preferred roads approximately 50% of the time, user 02 used the preferred roads 85% of the time and their trajectory included fewer road violations.

|  | | | Violated Specifications | | | |
|---|---|---|---|---|---|---|
| User | Inferred Expertise ($\hat{\beta}$) | Inferred Preferences ($\hat{\theta}$) | Path Length | Wrong Road Direction | Restricted Zones | Road Usage | Changes in Direction |
| 01 | 0.28 | [0.64, 0.36, 0.0] | 148 | 3 | 0 | 86/148 | 14 |
| 02 | 0.3 | [0.40, 0.40, 0.20] | 130 | 15 | 0 | 110/130 | 25 |
| 05 | 0.99 | [0.66, 0.0, 0.34] | 168 | 36 | 6 | 127/168 | 47 |
| 06 | 0.3 | [0.50, 0.50, 0.0] | 162 | 23 | 0 | 70/162 | 33 |
| 07 | 0.87 | [0.67, 0.06, 0.27] | 132 | 15 | 0 | 98/132 | 21 |
| 08 | 1.0 | [0.67, 0.0, 0.33] | 176 | 40 | 0 | 130/176 | 39 |

TABLE V: Predicted expertise level and preferences for user data. Preferences are presented in the following order: path length ($\theta_1$), avoidance of restricted areas ($\theta_2$), usage of preferred paths and travel directions ($\theta_3$). The last two columns indicate how many of the states visited during tele-operation correspond to the roads specified by the user and the number of times the user took actions that resulted in a change in direction.

|  | | Violated Specifications | | | |
|---|---|---|---|---|---|
| User | Path Length | Wrong Road Direction | Restricted Zones | Road Usage | Changes in Direction |
| 01 | 122 | 5 | 0 | 0/122 | 6 |
| 02 | 120 | 3 | 0 | 112/120 | 7 |
| 05 | 132 | 3 | 0 | 128/132 | 5 |
| 06 | 112 | 37 | 0 | 33/112 | 3 |
| 07 | 118 | 1 | 0 | 115/118 | 7 |
| 08 | 110 | 0 | 0 | 110/110 | 3 |

TABLE VI: Performance metrics for the optimal trajectories obtained using users' inferred preferences. The last two columns indicate how many of the states visited in the optimal trajectory correspond to the roads specified by the user and how many changes in direction were observed in the optimal trajectory.

The second group, with predicted expertise ($\hat{\beta} < 0.7$), corresponds to users that provided poor demonstrations of their intended preferences as determined by our proposed inference approach. On the one hand, we observed that although some users (i.e., users 05 and 08) seemly attributed a higher importance to reaching the goal state as quickly as possible, their demonstrated trajectories were the longest among all users. Additionally, these users also incurred approximately twice as many specification violations as the users in the first group. On the other hand, although user 07 demonstrated a trajectory closer in performance to the ones observed for the more expert users, our joint inference approach indicates that their demonstration seem to be poorly aligned with their inferred preferences. A closer look to the posterior distribution for this user shows ambiguity around two modes (i.e, high expertise with zero preference for road usage and low expertise with zero preference for avoidance of restricted areas), which in turn can explain the low expertise estimate despite the initially observed good performance.

*2) Comparison to High Expertise Trajectories:* To determine if the proposed inference approach was equally capable of inferring the users unknown preferences when good and poor demonstrations were provided, we compare the observed tele-operated and optimal trajectories obtained from the inferred preferences (i.e., $\beta = 0.001$). Results are shown for one user with high expertise (see Fig. 10) and one user with low expertise (see Fig. 11). From these figures, it can be observed that the optimal trajectories are similar to the ones demonstrated by the user, with the inclusion of fewer changes in direction and road violations. The same trajectory performance metrics considered in Sec. VIII-B1 were computed for the optimal trajectories and are listed in TABLE VI. Overall, we observed that the trajectories generated from the users' inferred preferences are shorter, include fewer changes in direction and for most cases incurred none or fewer specification violations. In particular, it was observed that the optimal trajectories obtained for the users that seemingly showed a preference for roads include road usage of approximately 97%. For users 01 and 06, an increase in the number of actions contrary to the specified road directions was observed in their corresponding optimal trajectory. This corresponds to their observed behaviour, and is explained by their inferred preferences, in which zero importance was attributed to road following.

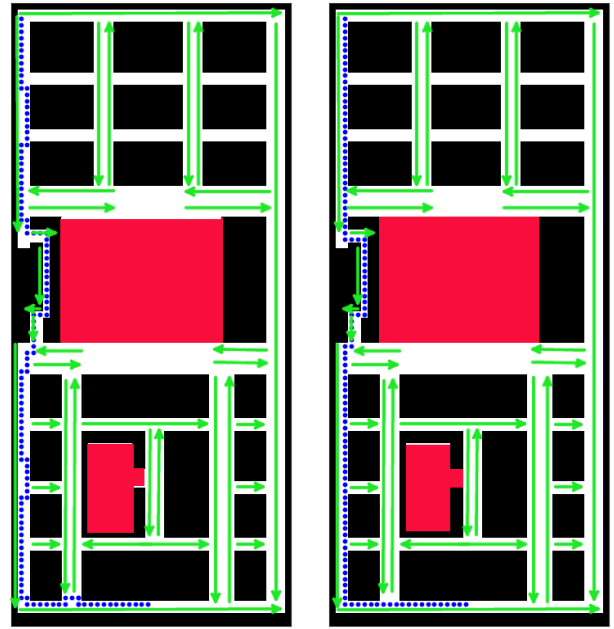

Fig. 10: Environment specification and tele-operated trajectory for a user (02) with higher expertise are shown in the left. The right image depicts the optimal trajectory for this user given their inferred preferences. Blue dots depict the trajectory the user followed from the top-left corner to one of three possible target locations in the opposite side of the environment. Green arrows indicate road specifications and red rectangles correspond to restricted areas.

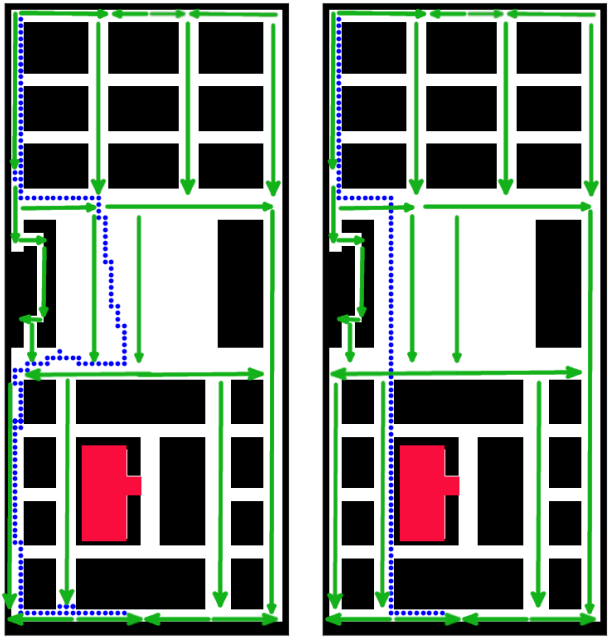

Fig. 11: Environment specification and tele-operated trajectory for a user with low expertise (08) are shown in the left. The right image depicts the optimal trajectory for this user given their inferred preferences. Blue dots depict the trajectory each user followed from the top-left corner to one of three possible target locations in the opposite side of the environment. Green arrows indicate road specifications and red rectangles correspond to restricted areas.

## IX. Discussion

Learning from the information conveyed by human actions is increasingly important as robotic systems enter human environments. However, existing approaches strongly rely on the assumption that demonstrations were generated by experts. In this work, we argue that when inferring information from humans' actions, robots should explicitly reason about the level of expertise of the human from which the observations were obtained. To explicitly reason about the human's expertise level, we propose to model a human's decision process as a Maximum Entropy policy in which the stochasticity coefficient $\beta$ approximates their true expertise level. Given this formulation, observed action sequences with low stochasticity (i.e., $\beta \rightarrow 0$) are associated with expert humans, while observations in which optimal actions are less frequently chosen (i.e., $\beta \rightarrow \infty$) are attributed to humans with lower expertise.

We presented a general formulation for estimating the proposed expertise coefficient $\beta$ along with the human preferences. This formulation was built on the assumption of a shared understanding about the task dynamics and state features of interest between the human and the robot. Two approaches, namely discrete-set and optimization-based, for the solution of this joint inference problem were proposed and tested in simulation. The evaluation in simulation showed that erroneous assumptions about a human's preferences or expertise level result in incorrect estimates when compared to the proposed formulation in which the robot jointly reasons about these two parameters. In particular, erroneous expertise assumptions during inference led to incorrect preference estimates. Similarly, expertise estimates obtained under erroneous preference assumptions resulted in mis-estimation of true expertise. Regarding the merits of these two proposed inferring approaches, on the one hand we showed that the discrete-set approach offers good performance, is faster to compute and requires fewer episodes, however its accuracy is dependent on the quality and diversity of the set of hypotheses considered during inference. On the other hand, although the optimization-based approach requires more observations and takes longer to compute, it can produce better estimates than the discrete set.

An application of the proposed formulation and discrete inference approach was illustrated on real user data. The data consisted of Cartesian trajectories users produced when tele-operating a robot through a warehouse environment for which they had previously defined a navigation specification. Using the discrete set inference approach, we estimated the users' preferences regarding the importance they attributed to reaching the target destination and following the environment specification they created for the robot. Overall, our results show that users who took longer to reach the goal and incurred a higher number of specification violations were predicted to have lower expertise compared to users whose tele-operation paths were shorter and with fewer violations. Finally, by comparing the observed demonstrations to the optimal trajectories obtained from the users' inferred preferences, we observed that the proposed joint inference approach was capable of inferring the users' preferences even when the users failed to properly demonstrate these preferences due to a lack of expertise.

There are several important limitations to our work. First, as shown in the evaluation of the discrete set inference approach, although the proposed joint inference formulation can successfully discriminate between the different expertise groups using few episodes, it requires a larger number of observations in order to disambiguate expertise coefficients within each group. This was in particular observed for the

$\beta$ values in the low expertise group. Second, both inference approaches work in a passive manner and attribute the same information value to all observed actions. This in turn results in a larger number of observations being required during inference. Third, our current observation model requires the computation of both $V^{\text{soft}}_{\theta,\beta}(\cdot)$ and $Q^{\text{soft}}_{\theta,\beta}(\cdot)$ for each candidate pair of preferences and expertise considered during inference. Although the discretization of the hypotheses space in the discrete set approach and the adoption of a grid-based sampling algorithm in the optimization approach were sufficient to validate the proposed joint inference formulation, a more efficient way of approximating the human observation model during inference is needed in order to scale our method to more complex and larger spaces. Lastly, our current formulation relies strongly on the assumption that the state features known to the robot fully capture the human's preferences. Although this structured formulation of the human's reward function facilitates inference, if not carefully designed, the chosen set of state features to consider during inference can potentially fail to capture and explain the observed human behavior. This in turn could also result in poor expertise estimates.

In future work we hope to address some of these limitations. In particular, we are interested in extending the current passive inference formulation into an active learning setting in which the robot can quickly gain information about the human preferences and expertise level by prompting the human's response at specific task states. Additionally, we want to explore ways in which the expertise and preferences of a human are not assumed to be stationary throughout a given task.